# Enhancing Attention with Higher-Order Attention

R. I. Cukier

*Department of Chemistry, Michigan State University, East Lansing, Michigan 48824-1322, USA*

cukier@chemistry.msu.edu

**Abstract**

Current decoder large language models rely on a combination of attention, feed forward and residual layers. The attention used is autoregressive (AR) though this is not strictly correct, as we note. That motivates the introduction of other attention methods that, while also not strictly AR, do enhance the AR character. One is based on a decomposition of mutual information (MI) into a set of Conditional MIs (CMIs) that correlate key tokens with pairs of query tokens. A second, higher-order attention, introduces correlations between contiguous key and query tokens. A third introduces correlations between a set of key and query tokens and uses a Generalized Maximum Entropy Principle to obtain an attention expression spanning a sequence of tokens. All these methods may enhance the information content of the autoregressive probabilities that are used in inference/generation. Applications to contrastive decoding and a systematic analysis of autoregressive probabilities are presented.

## 1. Introduction

Attention-based decoder large language Models (LLMs), with their autoregressive character, are currently the dominant method for training language models [1,2]. Each layer's module consists of attention (ATTN) and MLP/FFN contributions, both summed with a residual connection and layer normalization. As emphasized in e.g. Elhage et al. [3], ATTN mixes tokens in such a way that accounts for their sequence. The non-mixing MLP/FFN incorporates abstract token interactions not relying on strict sequence order.

In this work, our focus is on the (self) ATTN contribution to a decoder LLM

$$\tilde{x}_t := \sum_{s\le t} p\left(k_s | q_t\right) v_s \; \left(t = 1,2,\ldots T\right) \tag{1.1}$$

where, for a particular query $q_t$, the conditional probabilities $p\left(k_s|q_t\right)$ of the $k_s$ keys are summed over the $\left(s \le t\right)$ tokens as introduced by the value vectors, $v_s$. The queries, keys and

values are all from the same token set, $x_t \in X$, from a vocabulary $\mathcal{V}$ of size $|\mathcal{V}|$, in a $T-$ length context. The conditional probabilities (attention weights) $p\left(k_s | q_t\right)$ are obtained by optimization of

$$p\left(k_s | q_t\right) := \text{softmax}\left(\left[k_s^T \cdot q_t / \sqrt{d}\right]\right); \ \left(\sum_{s=1}^{s=T} p\left(k_s^T | q_t\right) = 1\left(t \in T\right)\right). \quad (1.2)$$

For $x_t = W_E \cdot w_t$ positionally embedded tokens, query, key and value vectors are respectively parameterized as: $q_t = W_Q \cdot x_t$; $k_t = W_K \cdot x_t$; $v_t = W_V \cdot x_t$ and output with $W_O \tilde{x}_t$. Neglecting biases, the scheme is parameterized using matrices: $W_E, W_Q, W_K, W_V, W_O, W_U$ denoting, respectively, embedding, query, key, value, output, unembedding matrices. (In the following we suppress the embed dimension $d$ -factor for convenience.)

The autoregressive character of the scheme is embodied in the linear, additive expression of Eq. (1.1). However, what is being approximated by Eq.(1.1) is the formal autoregressive decomposition [4] of the $T$ - random variable (RV) probability over the $T$ - length window:

$$P\left(x_1, x_2, \ldots, x_T\right) = \prod_{t=T}^{t=1} p\left(x_t | x_{t-1}, \ldots, x_1\right) := \prod_{t=T}^{t=1} p\left(x_t | x_{t-1:1}\right). \quad (1.3a)$$

The negative log-likelihood loss (NLL)

$$\mathcal{L} := -\log P\left(x_1, x_2, \ldots, x_T\right) = -\sum_{t=T}^{t=1} \log p\left(x_t | x_{t-1:1}\right) \quad (1.3b)$$

is the typical function optimized in training a LLM [5]. To try to connect with the standard attention expression in Eq. (1.1) make a Markov *Chain* (MC) approximation [6] to Eq. (1.3):

$$\prod_{t=T}^{t=1} p\left(x_t | x_{t-1:1}\right) \underset{MC}{\rightarrow} \prod_{t=T}^{t=1} p\left(x_t | x_{t-1}\right) := P_{MC}\left(x_1, x_2, \ldots, x_T\right). \quad (1.4a)$$

This is *not* a Markov *process* but rather a local approximation to the exact autoregressive decomposition of $P\left(x_1, x_2, \ldots, x_T\right)$. It is the lowest-level approximation to Eq. (1.3a) that keeps token correlations. If these pair conditionals were known, then the chain's probability could be sampled. In the LLM machine learning context,

$$\log P_{MC} = \sum_{t=T}^{t=1} \log p\left(x_t | x_{t-1}\right) \quad (1.4b)$$

could be optimized by NLL gradient descent. Clearly, this is not what is optimized by NLL gradient descent of the probability $p\left(x_t\right) := \text{sm}\left(\tilde{x}_t\right)\left(t \in T\right)$, with sm denoting the softmax function, obtained from Eq.(1.1). Thus, one must conclude that attention as currently formulated is, at best, "pseudo-autoregressive", and suggests that methods incorporating other correlation could be useful.

The idea is to introduce more correlation into the conditional or the unconditional parts of autoregressive probability to the training of a LLM. In Section 2, we do so via conditional mutual information and focus on $p\left(k_s | q_t, q_{T+1}\right)$ where $k_s$ are key and $q_t\left(t=1,2,...,T\right)$ and $q_{T+1}$ are query tokens. In Section 3 we introduce higher-order attention of the form $p\left(k_{s+1}, k_s | q_t\right)$ and further generalizations of the number of keys. In Section 4, we use trained probabilities $p\left(k_{s+s'}, k_s | q_t\right)$ $\left(s'=1,2,..,\tau\right)$ (with $\tau \ll T$ as a practical matter) along with a Generalized Maximum Entropy Procedure (GMEP) as a post-training phase to extend the autoregressive probability to a sequence of length $\tau$ using $p\left(x_{t+t'}, x_t | x_{<t}\right)_{t'\in\tau}$.

It is essential to realize for the attention versions relying on the $p\left(k_{s+s'}, k_s | q_t\right)$ that the identity

$$p\left(x_{t+t'}, x_t | x_{<t}\right) = p\left(x_{t+t'} | x_{<t+t'}\right) p\left(x_t | x_{<t}\right) \tag{1.5}$$

shows that there would be no difference with using the standard attention expressions on the right-hand side of the above versus that on the left-hand side. That would be the case under two conditions: 1) That training was exact and 2) that the NLL objective function did represent the linear combination over values of the attention method. With regard to item 1), the parameterization intrinsic to neural networks means that for the left and right sides of Eq. (1.5), one obtains, respectively, $p_\theta\left(x_{t+t'}, x_t | x_{<t}\right)$ and $p_\varphi\left(x_{t+t'} | x_{<t+t'}\right) p_\varphi\left(x_t | x_{<t}\right)$, with manifestly different parameterizations: $\theta$ and $\varphi$.

**2. Derivation of higher-order attention based on Conditional Mutual Information**

We summarize, and manipulate for our purposes, expressions from Cover and Thomas' "Elements of Information Theory" book [4]. The mutual information $I\left(X;Y\right)$ between RVs $X$ and $Y$ is defined in terms of entropy $H\left(X\right)$ and conditional entropy $H\left(X | Y\right)$ as

$$I\left(X;Y\right) = H\left(X\right) - H\left(X | Y\right). \tag{2.1}$$

Straightforward extension of Eq. (2.1) to the RV set $X_1,...,X_T,Y$ is

$$I(X_1,...,X_T,Y)=H(X_1,...,X_T)-H(X_1,...,X_T\,|\,Y). \tag{2.2}$$

Using the autoregressive probability decomposition in Eq. (1.3a), chain rules for entropy and conditional entropy follow immediately:

$$H(X_1,...,X_{T-1})=\sum\nolimits_{t=1}^{t=T-1}H(X_t\,|\,X_{t-1},...,X_1) \tag{2.3a}$$

and

$$H(X_1,...,X_{T-1}\,|\,Y)=\sum\nolimits_{t=1}^{t=T-1}H(X_t\,|\,X_{t-1},...,X_1,Y). \tag{2.3b}$$

Their difference

$$I(X_1,...,X_{T-1},Y)=\sum\nolimits_{t=1}^{t=T-1}\left[H(X_t\,|\,X_1,....,X_{t-1})-H(X_t\,|\,X_1,....,X_{t-1},Y)\right] \tag{2.4}$$

expresses the MI as a sum of conditional mutual information (CMI) terms

$$I(X_1,...,X_{T-1},Y)=\sum\nolimits_{t=1}^{t=T-1}I(X_t,Y\,|\,X_1,....,X_{t-1}):=\sum\nolimits_{t=1}^{t=T-1}I\left(X_t,Y\,|\,X_{t-1:1}\right). \tag{2.5}$$

It is a “chain rule for information” [4] relating the MI on left hand side to a sum of CMIs on the right-hand side of Eq. (2.5) and follows from the autoregressive decomposition in Eq. (1.3a).

These conditional MIs are “natural” extension of the CMI $I(X,Y\,|\,Z)$ to composite RVs $Z:=X_1,X_2,...,X_{t-1}$, with explicit form

$$I\left(X_t,Y\,|\,X_{t-1:1}\right)=\sum_{x_{t-1:1}}\sum_{x_t}\sum_{y}p\left(x_t,y,x_{t-1:1}\right)\log\left[\frac{p\left(x_t,y\,|\,x_{t-1:1}\right)}{p\left(x_t\,|\,x_{t-1:1}\right)p\left(y\,|\,x_{t-1:1}\right)}\right]. \tag{2.6}$$

Note that the expectation is over the distribution of all the RVs. These CMIs are the reduction in uncertainty in $X_t$ due to knowledge of $Y$ with the $X_{t-1:1}$ $(t=1,...,T-1)$ given. For purposes of optimization, decompose Eq. (2.6) as

$$I\left(X_t,Y\,|\,X_{t-1:1}\right)=\sum_{x_t,y}p(x_t,y)I_{x_t,y}\left(X_t,Y\,|\,X_{t-1:1}\right) \tag{2.7a}$$

with

$$I_{x_t,y}\left(X_t,Y\,|\,X_{t-1:1}\right):=\sum_{x_{t-1:1}}p\left(x_{t-1:1}\,|\,x_t,y\right)\log\left[\frac{p\left(x_{t-1:1}\,|\,x_t,y\right)}{p\left(x_{t-1:1}\,|\,x_t\right)p\left(x_{t-1:1}\,|\,y\right)}\right]. \tag{2.7b}$$

Note that the unconditional $p\left(x_t, x_{t-1:1}, y\right)$ aren't involved in the optimization. As a sensible choice for RV $Y$, set $Y = X_T$, the next token in sequence, to obtain $I\left(X_1,...,X_T\right)$.

To implement the above, as in the transition from Eq. (1.3b) to Eq. (1.1); that is, from negative log likelihood to attention, make the replacement $\sum_{x_{t-1:1}} ... \to \sum_{s=1}^{s=t-1} \sum_{x_s} ...$ to produce

$$I_{x_t,x_T}\left(X_t, X_T \mid X_{t-1:1}\right) \to \sum_{s=1}^{s=t-1} \sum_{x_s} I_{x_t,x_T}\left(X_t, X_T \mid X_s\right). \tag{2.8}$$

That leads to the replacement $p\left(x_{t-1:1} \mid x_t, x_T\right) \to p\left(x_s \mid x_t, x_T\right)$, along with the other two conditional probabilities in Eq. (2.7b). In Eq. (2.8), $x_{t-1:1} \to k_s$ represents each key token, and $q_T$ and $q_t \left(t = 1,2,...,T-1\right)$ are a pair of query tokens. To be minimized is the conditional mutual information (the Kullback-Lieber distance) between $p\left(k_s \mid q_t, q_T\right)$ and $p\left(k_s \mid q_t\right) p\left(k_s \mid q_T\right)$. The resulting parameterized $p_\theta\left(k_s \mid q_t, q_T\right)$'s are used to create an analogous to Eq. (1.1) attention-like expression:

$$\tilde{x}_{t,T} = \sum_{s \le t}^{s=T-1} p_\theta\left(k_s \mid q_t, q_T\right) v_s; \; \left(t = 1,2,...,T-1\right). \tag{2.9}$$

In this view, the probabilities of tokens (with $s \le T-1$) conditional on pairs of query tokens (between $t$ and $T-1$) are to be obtained in similar fashion to the construction of the conventional attention in Eq. (1.1). The $T-1$ added correlations are between query tokens separated as indicated in Eq. (2.9). This long-range correlation along the sequence may reinforce attention.

A parameterized expression for this attention expression is, via token embedding,

$$p\left(k_s \mid q_t, q_T\right) = \left[\sum_{s<T} e^{\mathbf{x}_s^T \cdot \mathbf{W}_{KQ} \cdot \left(\mathbf{x}_t + \mathbf{x}_T\right)}\right]^{-1} e^{\mathbf{x}_s^T \cdot \mathbf{W}_{KQ} \cdot \left(\mathbf{x}_t + \mathbf{x}_T\right)}$$

$$:= \text{softmax}\left(\mathbf{x}_s^T \cdot \mathbf{W}_{KQ} \cdot \left(\mathbf{x}_t + \mathbf{x}_T\right)\right); \; \mathbf{W}_{KQ} := \mathbf{W}_K^T \cdot \mathbf{W}_Q \tag{2.10}$$

$$\text{s.t.} \; \sum_{s<T} p\left(k_s \mid q_t, q_T\right) = 1; \; \left(t = 1,2,...,T-1\right).$$

**3. Derivation of higher-order attention based on local Markov chains**

An alternative approach to enhancing attention starts from the formal decomposition in Eq. (1.3a) that we now write for composite RVs. To that end, partition integer sequence $(n, n-1, \ldots, 1) := (n:1)$ into $r-1$ slices with $k = n/r$ descending-ordered consecutive integers $n_r$ per slice. Let $\bar{n}_r$ indicate the remaining descending-ordered integers to that $n_r$. Then (excluding the last, singlet probability term)

$$P(n:1) = p(n_{r-1} | \bar{n}_{r-1}) \, p(n_{r-2} | \bar{n}_{r-2}) \ldots p(n_1 | \bar{n}_1) \tag{3.1}$$

with the corresponding Markov chain approximation

$$P_{MC}(n:1) = p(n_{r-1} | n_{r-2}) \, p(n_{r-2} | n_{r-3}) \ldots p(n_1 | n_0) \;\; (n_0 := \bar{n}_1). \tag{3.2}$$

For example: $r = 5; k = 3; n = kr = 15$ has 4 slices

$$P(15:1) = p(15,14,13 | 12:1) \, p(12,11,10 | 9:1) \, p(9,8,7 | 6:1) \, p(6,5,4 | 3:1) \tag{3.3}$$

and

$$P_{MC}(15:1) = p(15,14,13 | 12:10) \, p(12,11,10 | 9:7) \, p(9,8,7 | 6:4) \, p(6,5,4 | 3:1). \tag{3.4}$$

For example: $r = 6; k = 2; n = kr = 12$ has 5 slices

$$P(12:1) = p(12,11 | 10:1) \, p(10,9 | 8:1) \, p(8,7 | 6:1) \, p(6,5 | 4:1) \, p(4,3 | 2:1) \tag{3.5}$$

and

$$P_{MC}(12:1) = p(12,11 | 10,9) \, p(10,9 | 8,7) \, p(8,7 | 6,5) \, p(6,5 | 4,3) \, p(4,3 | 2,1). \tag{3.6}$$

It would be impractical to deal with long Markov chains, as each additional conditional probability requires a partition function as a normalizing factor in the softmax construction. Thus, apply this MC idea at a low order to build higher-order attention. Namely,

$$P_{MC}(n:1) = p(n_2 | n_1) \, p(n_1 | n_0) \to p\big(k(T_2) | k(T_1)\big) P\big(k(T_1) | q(T_0)\big). \tag{3.7}$$

This latter expression symbolizes the number of keys $k(T_i)\,(i = 1,2)$ and queries $q(T_0)$ that will be needed to produce attention with increasing degrees of token correlation. Thus, at lowest MC order,

$$\tilde{x}_t = \sum\nolimits_{s \le t} p(k_s | q_t) \, v_s \;\; (t = 1,2,\ldots,T) \text{ (STD)} \tag{3.8}$$

where STD denotes the standard attention [5].

With the assumption that $\tilde{x}_t \to p(x_t \mid x_{<t})$, the implicit autoregressive STD autoregressive attention expression introduces new correlations between keys:

$$\tilde{x}_t = \sum_{s+1\le t} p(k_{s+1} \mid k_s)\, p(k_s \mid q_t)\left[v_s + v_{s+1}\right](t = 1,2,\ldots,T) \text{ (singlet MC)} \tag{3.9}$$

where the singlet MC introduces neighbor key correlations separated from the query tokens per the attention paradigm. Note that the singlet MC can also be formulated via

$$\tilde{x}_t = \sum_{s+1\le t} p(k_{s+1}, k_s \mid q_t)\left[v_s + v_{s+1}\right](t = 1,2,\ldots,T) \text{ (singlet)} \tag{3.10}$$

as follows from the autoregressive identity $p(x_3 x_2 \mid x_1) = p(x_3 \mid x_2 x_1)\, p(x_2 \mid x_1)$ followed by the MC $p(x_3 x_2 \mid x_1) \underset{MC}{\to} p(x_3 \mid x_2)\, p(x_2 \mid x_1)$. The singlet requires a higher-order attention-like term, $p(k_{s+1}, k_s \mid q_t)$. The $\tilde{x}_t$ obtained from Eq. (3.9) is a representation of that from Eq. (3.10). With the attention paradigm, both versions of $\tilde{x}_t$ imply, with Eq. (1.5), that $p(x_t, x_{t+1} \mid x_{<t}) := \text{sm}(\tilde{x}_{t+1})\text{sm}(\tilde{x}_t)$, with their incorporation of neighbor-token correlation. The downside of the, in principle, more accurate singlet expression in Eq. (3.10) is that it scales with $|\mathcal{V}|^3$ versus the $|\mathcal{V}|^2$ of the singlet MC in Eq. (3.9).

Proceeding systematically to higher order in key token correlation

$$\tilde{x}_t = \sum_{s+2\le t} p(k_{s+2} \mid k_{s+1})\, p(k_{s+1} \mid k_s)\, p(k_s \mid q_t)\left[v_s + v_{s+1} + v_{s+2}\right] \text{ (doublet MC)} \tag{3.11}$$

The scaling is $3|\mathcal{V}|^2$ per query, thus still of the complexity of conventional attention. The scheme can be extended to triplet MC, and beyond. Note that, owing to the multiplicative MC form, back propagation is done with each term's gradient multiplied by the remaining probabilities.

The requisite probabilities can be parameterized in "multiplicative" and "additive" versions.

The multiplicative versions are:

$$\begin{aligned}
&p(k_s \mid q_t) := \text{softmax}\left(\left[\mathbf{x}_s^T \cdot \mathbf{W}_{KQ} \cdot \mathbf{x}_t\right]\right) \\
&p(k_{s+1}, k_s \mid q_t) := \text{softmax}\left(\left[\mathbf{x}_{s+1}^T \cdot \mathbf{W}_{KK} \cdot \mathbf{x}_s\right]\left[\mathbf{x}_t^T \cdot \mathbf{W}_{Q1} \cdot \mathbf{1}^T\right]\right) \\
&\text{with} \\
&\mathbf{W}_{K^+K^+} := \mathbf{W}_{K^+}^T \cdot \mathbf{W}_{K^+};\ \mathbf{W}_{KK} := \mathbf{W}_K^T \cdot \mathbf{W}_K;\ \mathbf{W}_{KQ} := \mathbf{W}_K^T \cdot \mathbf{W}_Q;\ \mathbf{W}_{Q1} := \mathbf{W}_Q^T \cdot \mathbf{W}_1 .
\end{aligned} \tag{3.12}$$

(For the $p(k_{s+1}, k_s \mid q_t)$ term, we introduce a new, learnable matrix $\mathbf{W}_1$ and a model-dimensioned column vector $\mathbf{1}^T$ ).

The additive versions are:

$$\begin{aligned}
&p(k_s \mid q_t) := \text{softmax}\left(\mathbf{x}_s^T \cdot \mathbf{W}_{KQ} \cdot \mathbf{x}_t\right) \\
&p(k_{s+1}, k_s \mid q_t) := \text{softmax}\left(\left(\mathbf{x}_{s+1}^T + \mathbf{x}_s^T\right) \cdot \mathbf{W}_{KQ} \cdot \mathbf{x}_t\right) \\
&\text{with } \mathbf{W}_{KQ} := \mathbf{W}_K^T \cdot \mathbf{W}_Q .
\end{aligned} \tag{3.13}$$

The different information contents of the underlying determinants of attention are readily obtained from the following (manifestly positive) mutual information:

$$I(k_{s+1}, k_s \mid q_t) = \mathbb{E}^{\theta}_{(k_{s+1}, k_s \mid q_t)} \log\left[\frac{p_\theta(k_{s+1}, k_s \mid q_t)}{p_\theta(k_{s+1} \mid q_t)\, p_\theta(k_s \mid q_t)}\right]. \tag{3.14}$$

Once the $\theta$ parameters of $p_\theta(k_{s+1}, k_s \mid q_t)$ are obtained, the above denominator can be back calculated using Eq. (3.13), putting the ingredients of this mutual information onto a consistent basis. In Eq. (3.14) we left the expectation conditional on the query token $q_t$ to allow investigation of the information content conditional on a query token.

## 4. Increasing the attention order with Generalized MEP (GMEP)

In order to increase $\tau$ , the length of the token sequence considered, a Maximum Entropy Principle (MEP), introduced to Statistical Mechanics by Jaynes [7-9], can be invoked. The MEP maximizes entropy with suitable constraints on the various probabilities enforced by Lagrange Multipliers (LMs). MEP procedures were introduced to sequence-based data [10], and generalized in various ways [11]. One of them, the generalized MEP (GMEP) , can be tailored to LLM inference.

First, train the model to obtain the $p\left(k_{s+s'},k_s \mid q_t\right)$ $\left(s'=1,2,..,\tau\right)$ (with $\tau \ll T$ ) and use the attention mechanism in Eq. (3.10) along with Eq. (1.5) to obtain $p\left(x_t,x_{t+t'} \mid x_{<t}\right) := \mathrm{sm}\left(\tilde{x}_t\right)\mathrm{sm}\left(\tilde{x}_{t+t'}\right)$ for token pairs $\left\{x_t, x_{t+t'}\right\}$ with $\left(1 \le t < t' \le t+\tau\right)$. While requiring additional calculation relative to $s'=1$ in Eq. (3.10), the $\tau's$ under consideration can't be too large (e.g. $\tau < 5$) as their use for even e.g. Look Ahead Greedy Search (L.A.G.S.) would become too expensive for large $\tau$ .

Thus, do GMEP on $p\left(x_t,x_{t+t'} \mid x_{<t}\right)$ to obtain $\left\{p\left(x_t,x_{t+t'} \mid x_{<t}\right)\right\}_{t'\in\tau} \underset{\text{GMEP}}{\rightarrow} p\left(x_{t+\tau:t} \mid x_{<t}\right)$ over sequences of length $\tau$ , as follows. The constraint equation on the target distribution $f$ , for token pair $\left\{x_t,x_{t'}\right\}$ is

$$\sum_{\mathrm{X}_-} f\left(\mathrm{X}\right) - p\left(x_t,x_{t'} \mid x_{<t}\right) = 0 \;\; \left(1 \le t < t' \le t+\tau\right),$$

$$\text{with } \mathrm{X} = \left(x_1,x_2,...,x_\tau\right) \text{ and } \mathrm{X}_- \doteq \mathrm{X} - \left\{x_t,x_{t'}(t<t')\right\}. \tag{4.1}$$

The Lagrangian is

$$L = -\sum_{\mathrm{X}} f\left(\mathrm{X}\right)\log f\left(\mathrm{X}\right) + \sum_{x_2,x_1} \lambda_{x_2,x_1}\left[\sum_{\mathrm{X}_-} f\left(\mathrm{X}\right) - p\left(x_2,x_1 \mid x_{1:1}\right)\right] + \sum_{x_3,x_1} \lambda_{x_3,x_1}\left[\sum_{\mathrm{X}_-} f\left(\mathrm{X}\right) - p\left(x_3,x_1 \mid x_{1:1}\right)\right] + \ldots + \sum_{x_\tau,x_{\tau-1}} \lambda_{x_\tau,x_{\tau-1}}\left[\sum_{\mathrm{X}_-} f\left(\mathrm{X}\right) - p\left(x_\tau,x_{\tau-1} \mid x_{\tau-1:1}\right)\right], \tag{4.2}$$

with Lagrange multipliers: $\lambda_{x_t,x_{t'}}$ and $x_{t:1} := x_t,x_{t-1},...,x_1$.

The Euler-Lagrange equation is

$$-1-\log f\left(\mathrm{X}\right) + \lambda_{x_1,x_2} + \lambda_{x_1,x_3} + \ldots + \lambda_{x_{\tau-1},x_\tau} = 0. \tag{4.3}$$

Noting that $S := \sum_{t=1}^{\tau-1}\sum_{t'=t+1}^{\tau} \lambda_{x_t,x_{t'}}$ will generate the above-sum of Lagrange Multiplier terms, the Euler-Lagrange equation can be compacted to

$$-1-\log f\left(\mathrm{X}\right) + S = 0. \tag{4.4}$$

The scaling of this MEP follows from the $(\tau-1)\tau/2*|\mathcal{V}|^2 \sim \tau^2/2*|\mathcal{V}|^2$ Lagrange Multipliers to be determined for the vocabulary $\mathcal{V}$. With a candidate set $\mathcal{V}_C$ (see below for a method to choose it), such that $|\mathcal{V}_C| \ll |\mathcal{V}|$, the scaling is order $\tau^2/2*|\mathcal{V}_C|^2$. Optimization schemes such as the well-known Newton-Krylov method [12] are appropriate and, for large-scale systems, there are sophisticated methods available [13]. Two approaches that focus on the structure of the probability distributions can be used to speed up the calculation. Using PCA/SVD to obtain $\tau_C^2$ modes reduces the $\tau^2$ part to $\tau_C^2 \ll \tau^2$. Using Rényi entropy (with $\alpha > 1$), versus Gibbs-Shannon entropy, reduces the $|\mathcal{V}_C|$ value by further focusing on (the desired) higher probability parts of the distribution.

Once the Lagrange Multipliers are obtained, the optimized values in Eq. (4.4)

$$p\left(x_{t+\tau:t} \mid x_{<t}\right) = f^*\left(x_{t+\tau:t} \mid x_{<t}\right) \tag{4.5}$$

are the desired order-$\tau$ token sequence probabilities. Of course, compared to training, this operation is done only once.

The Eq. (4.5) probabilities can be used for inference/generation. Over some context $T$

$$p(x_{T:1}) := p(x_T, x_{T-1}, \ldots, x_1) = \sum_{t=[T/\tau]}^{t=1} p\left(x_{t\tau:(t-1)\tau+1} \mid x_{<(t-1)\tau+1}\right). \tag{4.6}$$

In the LM solution there are redundancies to eliminate. The pair probabilities satisfy the row-based (the last row is fixed) conditions:

$$p(x_\tau, x_{t'}) = p_{t'}(x_{t'}) - \sum_{t=1}^{\tau-1} p(x_t, x_{t'}) \left(1 \le t < t' \le \tau\right). \tag{4.7a}$$

Alternatively, they can be formulated as column-based (the last column is fixed)

$$p(x_t, x_\tau) = p(x_t) - \sum_{t'=1}^{\tau-1} p(x_t, x_{t'}) \left(1 \le t < t' \le \tau\right). \tag{4.7b}$$

When decoding, as a practical matter, a scheme most likely must be used to restrict the state space to a manageable size. Section 6 provides the series of inequalities:

$$\min_x p(X = x) \ge \min_x p(X = x \mid x_1) \ge \min_x p(X = x \mid x_2, x_1) \ge \ldots \ge \min_x p(X = x \mid x_{\tau:1}) \tag{4.8}$$

for any token $x_t \in X$ from the vocabulary $\mathcal{V}$. The $\min_x p(x)$ is an upper bound to all the conditional probabilities. Thus, pick the cut-off, candidate set $\mathcal{V}_c$ of $\mathcal{V}$ using $\min_x p(X = x)$ – i.e., just from the vocabulary. Then, for an e.g., L.A.G.S. of length $\tau$, the decoding scaling is $\sim |\mathcal{V}|_{\text{cut}}\, \tau$. That can turn an impractical search into a practical search for both higher-order attention and, especially, the GMEP scheme.

**5. Application to Contrastive Decoding**

The different methods presented in Sections 2-4 are quite natural for Contrastive Decoding (CD). [14] CD invokes two or more models to be used at inference/generation where one is considered the Expert (E) and the other(s) the amateur(s)/anti-expert(s) (A,…). The approach is based on a score $S(v_t)$ for the $v$'th token at position $t$ in a sequence:

$$S(v_t) \propto \log p_E(v_t \mid x_{<t}) - \alpha \log p_A(v_t \mid x_{<t}) \tag{5.1}$$

with $\alpha$ an adjustable parameter.

For any $t$, the score is normed to the CD probability:

$$p_{CD}(v) = \left[ p_E(v) / p_A^{\alpha}(v) \right] \Big/ \sum\nolimits_w \left[ p_E(w) / p_A^{\alpha}(w) \right]. \tag{5.2}$$

A token with high-to-moderate E and low A probability boosts the CD-based probability relative to its E value, while high-to-moderate E and high A probability suppresses the CD-based probability relative to its E value. Note that this scheme is based on the E probabilities; thus, they are in the high to middling range, while the corresponding A probabilities can range from high to low.

CD can lead to a reduction in repetitive text and reduce hallucinations, [15-17] and improve common sense reasoning [18]. Operating at inference/generation versus training is much less expensive than model development. However, when based on ATTN, it is intrinsically local to each position in a sequence versus optimized over a sequence. Our GMEP method, with its focus on sequence, may improve the quality of CD. To implement it for sequences, suppose the expert is taken from GMEP and the anti-expert from ATTN.

To generate the sequences for the E and A models in the simplest application consider using $\tau$ – length L.A.G.S. for the A model:

$$x_t^*(\mathsf{A}) = \arg\max_{x_t \in \mathcal{V}} \sum_{i=t}^{i=t+\tau} \log p\left(x_i \mid x_{<i}\right) \tag{5.3}$$

and use the GMEP procedure for the E model.

$$x_t^*(\mathsf{E}) = \arg\max_{t+\tau:t \in \mathcal{V}} \log p\left(x_{t+\tau:t} \mid x_{<t}\right) \ . \tag{5.4}$$

Again, as a practical matter, use the $\sim |\mathcal{V}|_{\text{cut}}^{\tau-1} |\mathcal{V}|$ restriction of the $p_E(v)$ to a high probability set. Using these respective probabilities, for the E and A models, will reorder the token probabilities from those of $p_E(v)$ to those of $p_{CD}(v)$ according to Eq. (5.2). The scheme can be applied to nucleus (top-p) and beam search sampling methods at the expense of more overhead to evaluate and store candidates.

The scheme could be elaborated with E from the GMEP and two anti-experts: A1 H.O. ATTN and A2 ATTN.

**6. Some implications of higher-order attention**

As the degree of autoregression decreases, the corresponding entropy should increase owing to the greater freedom of probabilities with decreased autoregression. To show this property, write a series of approximations to the exact $\tau$ - point probability

$$P_{\text{approx}}(\eta) := \prod_{t=T}^{t=1} p\left(x_t \mid x_{t-1:\eta}\right) \ \left(\eta = 2,3,\ldots,t-1\right) \tag{6.1}$$

where $P\left(x_1, x_2, \ldots, x_T\right) = P_{\text{exact}} = P_{\text{approx}}(\eta = 1)$ by Eq. (1.3a).

The corresponding entropies are

$$H_{\text{approx}}(\eta) = \sum_{t=T}^{t=1} H\left(X_t \mid X_{t-1:\eta}\right) \ \left(\eta = 2,3,\ldots,t-1\right); H_{\text{exact}} = H_{\text{approx}}(1) \,. \tag{6.2}$$

The conditional entropies $H\left(X_t \mid X_{t-1:\eta}\right)$ satisfy the inequalities [10]

$$H\left(X_t \mid X_{t-1:\eta}\right) \le H\left(X_t \mid X_{t-1:\eta+1}\right), \tag{6.3}$$

as can be obtained from Jensen's inequality applied to the convex entropy. Therefore, with Eq.(6.3), the approximate entropy in Eq.(6.2) satisfies

$$H_{\text{approx}}(\eta) \le H_{\text{approx}}(\eta+1) \,\square\,. \tag{6.4}$$

Thus

$$H_{\text{exact}} = H_{\text{approx}}(1) \le H_{\text{approx}}(2) \le \ldots \le H_{\text{approx}}(t-1), \tag{6.5}$$

with the last term in the chain of inequalities the independent random variable limit.

The above is an average statement, of course. A stronger, more interesting, result can be proved, as shown in Appendix A. Namely, as more conditional RVs in Eq. (6.1) are introduced, the difference between minimum and maximum probabilities increases, leading to the series of inequalities for the quantities defined in Eq. (6.1):

$$\begin{aligned} &\min P_{\text{exact}} = \min P_{\text{approx}}(1) \le \min P_{\text{approx}}(2) \le \ldots \le \min P_{\text{approx}}(t-1) \\ &\le \max P_{\text{approx}}(t-1) \le \ldots \le \max P_{\text{approx}}(2) \le \max P_{\text{approx}}(1) = \max P_{\text{exact}} \,. \end{aligned} \tag{6.6}$$

Increasing the conditionality of probabilities produces a greater spread in the min/max probability values. The min inequalities are particularly useful for systematically obtaining the probability cutoff $\mathcal{V}_c$ of the vocabulary $\mathcal{V}$.

Now focus on the maximum conditional probabilities, as they are of greatest interest. The context-$T$ probability of Eq. (1.3a) using the ATTN mechanism relies on the conditional decomposition where the conditional terms are only known implicitly. Thus, to investigate the maximum inequalities of Eq. (6.6) a roundabout procedure is needed. Suppose you were to train on shorter, truncated contexts indexed by $\eta = 2,3,\ldots,t-1$

$$P\left(x_T, x_{T-1},\ldots,x_\eta\right) := \prod_{t=T}^{t=\eta} p\left(x_t \mid x_{t-1:t-\eta}\right) \; \left(\eta = 2,3,\ldots,t-1\right)\left(x_0 := 0\right). \tag{6.7}$$

The corresponding conditionals are

$$p\left(x_t \mid x_{t-1:t-\eta}\right) := p\left(x_t, x_{t-1:t-\eta}\right) \big/ p\left(x_{t-1:t-\eta}\right) \tag{6.8}$$

and can be obtained by marginalization:

$$p\left(x_{t-1:t-\eta}\right) = \sum_{x_t} p\left(x_{t:t-\eta}\right). \tag{6.9}$$

However, as a practical matter, a method of evaluation of these high dimensional probabilities, such as top-K or its simplest, L.A.G.S., that provides an approximation to the maximum probability must be used. Designate the maximum conditioned probabilities using 'max' and those obtained by Greedy Search as 'K=1' (from top-K search).

Then it is evident, denoting $X_t = x_t^*$ as the maximizing value of RVs $\{X_t\}$, that

$$p^{\max}\left(x^{*}_{t-1:t-\eta}\right)=p^{K=1}\left(x^{*}_{t:t-\eta}\right)+\sum_{x_t\neq x_t^{*}}p\left(x_t,x^{*}_{t-1:t-\eta}\right). \tag{6.10}$$

Thus:

$$p^{\max}\left(x^{*}_{t-1:t-\eta}\right)\geq p^{K=1}\left(x^{*}_{t-1:t-\eta}\right), \tag{6.11}$$

and

$$p\left(x_t\mid x_{t-1:t-\eta}\right)=p\left(x_t,x_{t-1:t-\eta}\right)\big/p\left(x_{t-1:t-\eta}\right) \tag{6.12}$$

satisfies

$$p^{\max}\left(x_t\mid x_{t-1:t-\eta}\right)\leq p^{K=1}\left(x_t\mid x_{t-1:t-\eta}\right). \tag{6.13}$$

Eqs. (6.13) and (6.6) lead to the "pairwise" comparisons

$$p^{K=1}\left(x_t\mid x_{t-1:t-\eta}\right)\geq p^{\max}\left(x_t\mid x_{t-1:t-\eta}\right)\geq p^{\max}\left(x_t\mid x_{t-1:t-\eta+1}\right). \tag{6.14}$$

But the 'chain' of inequalities of Eq. (6.6) would be broken by (6.13). For example:

$$p^{K=1}\left(x_3\mid x_2,x_1\right)\geq p^{\max}\left(x_3\mid x_2,x_1\right)\geq p^{\max}\left(x_3\mid x_2\right)\leq p^{K=1}\left(x_3\mid x_2\right)\geq p^{\max}\left(x_3\mid x_2\right). \tag{6.15}$$

However, by definition, the L.A.G.S. approximation in fact sets the conditionals equal:

$$p^{K=1}\left(x_t\mid x_{t-1:t-\eta}\right)=p^{\max}\left(x_t\mid x_{t-1:t-\eta}\right). \tag{6.16}$$

Thus, to this degree of approximation, which is driven by practicality, the chain of inequalities for the maxima in Eq. (6.6) hold equally well when using L.A.G.S..

As a convenient measure of the degree of loss of information with shortening the conditional ranges in Eq. (6.6) it is convenient to use (maximum) perplexity

$$PP^{\max}\left(\eta\right):=\prod_{t=T}^{t=\eta}\frac{1}{\left(P^{\max}\left(x_t\mid x_{t-1:t-\eta}\right)\right)^{\frac{1}{T}}} \tag{6.17}$$

or the negative of its log:

$$\log PP^{\max}\left(\eta\right):=-\frac{1}{T}\sum_{t=T}^{t=\eta}\log p^{\max}\left(x_t\mid x_{t-1:t-\eta}\right). \tag{6.18}$$

This produces a chain of inequalities:

$$PP^{\max}_{\text{exact}}=PP^{\max}_{\text{approx}}\left(1\right)\leq PP^{\max}_{\text{approx}}\left(2\right)\leq\ldots\leq PP^{\max}_{\text{approx}}\left(T-1\right). \tag{6.19}$$

A decrease in autoregression increases perplexity.

## 7. Concluding Remarks

There is a contradiction between the conventional attention expression with its linear combination of values, conditional on each query's probability, and the autoregressive negative log-likelihood objective function that it is supposed to represent. They are not relatable by systematic approximation. The higher-order attention expressions introduced here are more correlated approximations to the NLL expression. Embedding more language structure in the training of the autoregressive probabilities that are then chained together for inference/generation may improve the coherence of the LLM output. Further, the empiricism inherent to training $p_\theta\left(x_{t+1}, x_t \mid x_{<t}\right)$, or its MC approximation, with parameters $\theta$, vs. training $p_\varphi\left(x_t \mid x_{<t}\right)$, with parameters $\varphi$, should provide a more accurate data representation in the former versus the latter. The various higher-order attentions developed here should produce systematically more specific probabilities with a richer account of language, though they may be impractical unless used for a restricted $\mathcal{V}$ alphabet.

The first alternative attention in Sec. 2 is based on the CMI decomposition of the mutual information that follows from the fundamental probability decomposition underlying the NLL objective. It, too, uses a linear combination of values but incorporates other correlation via its dependence on query token pairs separated in sequence by all token pairs, as displayed in Eq. (2.9). The second alternative of Sec. 3 is based on incorporating various levels of neighbor key token correlation. The third method in Sec. 4 does require the expense of training with pairs of keys over a $\tau$ interval, and the overhead of running post-training GMEP. But it may provide a much more informative version of attention via $p\left(x_{t+\tau:t} \mid x_{<t}\right)$.

Section 5 provides an application of higher-order attention to Contrastive Decoding. Sequence-based decoding can also provide an entrée to enhanced L.A.G.S. methods for use when resources are limited. Also, distillation [18], as well as other teacher-student focused methods, could profit from this sequence version of attention, as it may reduce exposure bias.

Section 6 presents a systematic exploration of the information content of the maxima of conditional probabilities/perplexity dependence on the extent of auto-regression. How the accuracy of next token prediction degrades with a decrease in autoregression may help in evaluating the quality of language models. referral

**Appendix A**

In this Appendix we prove that:

$$\min_{x_3,x_2,x_1} p(x_3 \mid x_2, x_1) \le \min_{x_3,x_2} p(x_3 \mid x_2) \le \max_{x_3,x_2} p(x_3 \mid x_2) \le \max_{x_3,x_2,x_1} p(x_3 \mid x_2, x_1) \quad \text{(A.1)}$$

and later straightforwardly generalize it to obtain the result underlying Eq. (6.6).

The proof relies on the law of iterated expectations:

$$\begin{aligned} &E(X) = E_Y E_{X|Y} = \int dy\, p(y) \int dx\, x\, p(x \mid y)\, dx = \int dy\, p(y) F(y) \\ &F(Y=y) := \int dx\, x\, p(x \mid Y=y); \quad E_Y F(Y) = \int dy\, p(y) F(y); \end{aligned} \quad \text{(A.2)}$$

meaning the $Y$ expectation is of the function $F(Y)$. Also of use is an indicator function: If $A \in X$ then $1_A(x) = 1$ if $x \in A$ and $1_A(x) = 0$ otherwise. Applying the law of iterated expectation from above, with $X \to X_3; Y \to X_1$

$$p(x_3 \mid x_2) = E_{X_3}\left[1_{X_3=x_3} \middle| X_2 = x_2\right] = E_{X_1}\left[E_{X_3}\left[1_{X_3=x_3} \mid X_2 = x_2, X_1\right] \middle| X_2 = x_2\right] \quad \text{(A.3)}$$

where $p(x_3 \mid x_2, X_1) := E_{X_3}\left[1_{X_3=x_3} \mid X_2 = x_2, X_1\right]$. Therefore

$$p(x_3 \mid x_2) = E_{X_1}\left[p(x_3 \mid x_2, X_1) \middle| X_2 = x_2\right]. \quad \text{(A.4)}$$

Thus, $p(x_3 \mid x_2)$ can be thought of as an average of $p(x_3 \mid x_2, \mathrm{X}_1)$ over $\mathrm{X}_1$, given fixed $x_2$ and $x_3$, implying that $p(x_3 \mid x_2)$ is less than the maximum and greater than the minimum of $p(x_3 \mid x_2, X_1)$ over $X_1$, given fixed $x_2$ and $x_3$.

To prove this supposition, focus on RV $X_1$. Given any fixed $x_2$ and $x_3$

$$\min_s p(x_3 \mid x_2, X_1 = s) \le p(x_3 \mid x_2, X_1) \le \max_s p(x_3 \mid x_2, X_1 = s) \quad \text{(A.5)}$$

by definition of minimum and maximum.

$$\begin{aligned} &\therefore \min_s p(x_3 \mid x_2, X_1 = s) \le E_{X_1}\left[p(x_3 \mid x_2, X_1) \middle| X_2 = x_2\right] \le \max_s p(x_3 \mid x_2, X_1 = s) \\ &\therefore \min_s p(x_3 \mid x_2, X_1 = s) \le p(x_3 \mid x_2) \le \max_s p(x_3 \mid x_2, X_1 = s), \text{ by (A.5)}. \end{aligned} \quad \text{(A.6)}$$

Thus,

$$\min_{s,t,u} p(X_3 = u \mid X_2 = t, X_1 = s) \le \min_{s} p(x_3 \mid x_2, X_1 = s) \le p(x_3 \mid x_2) \tag{A.7}$$

with the first inequality by definition of minimum and

$$p(x_3 \mid x_2) \le \max_{s} p(x_3 \mid x_2, X_1 = s) \le \max_{s,t,u} p(X_3 = u \mid X_2 = t, X_1 = s) \tag{A.8}$$

with the second inequality by definition of maximum. (Note that above is first for $s$ and then $s,t,u$ , so $s$ is same in both, and fixes $t,u$ .) Therefore

$$\min_{s,t,u} p(X_3 = u \mid X_2 = t, X_1 = s) \le p(x_3 \mid x_2) \le \max_{s,t,u} p(X_3 = u \mid X_2 = t, X_1 = s) \quad \square \tag{A.9}$$

Note that $\min_{s,t,u} p(X_3 = u \mid X_2 = t, X_1 = s)$ is a constant and is a lower bound on $p(x_3 \mid x_2)$ and that $\max_{s,t,u} p(X_3 = u \mid X_2 = t, X_1 = s)$ is a constant and an upper bound of $p(x_3 \mid x_2)$.

Therefore,

$$\begin{aligned} &\min_{s,t,u} p(X_3 = u \mid X_2 = t, X_1 = s) \le \min_{u,t} p(X_3 = u \mid X_2 = t) \\ &\le \max_{u,t} p(X_3 = u \mid X_2 = t) \le \max_{s,t,u} p(X_3 = u \mid X_2 = t, X_1 = s) \quad \square \end{aligned} \tag{A.10}$$

Thus the $X_1$ conditionals lower and upper bound the unconditional on $X_1$ probability, verifying Eq. (A.1).

To apply this result to $P_{\text{approx}}(\eta) := \prod_{t=1}^{t=T} p(x_t \mid x_{t-1:\eta})$ $(\eta = 2,3,\ldots,t-1)$ of Eq.(6.1), make the variable 'mapping':

$$\begin{aligned} &x_3 \to x_t;\ X_3 \to X_t = u \\ &x_2 \to x_{t-1}, x_{t-2}, \ldots, x_{\eta+1} := x_{t-1:\eta+1}; \quad X_{t-1:\eta+1} = w_{t-1:\eta+1} \\ &x_1 \to x_\eta;\ X_1 \to X_\eta = s\,. \end{aligned} \tag{A.11}$$

From Eq. (A.10) and Eq. (A.1),

$$\begin{aligned} &\min_{x_t, x_{t-1:\eta}} p(x_t \mid x_{t-1:\eta}) \le \min_{x_t, x_{t-1:\eta+1}} p(x_t \mid x_{t-1:\eta+1}) \\ &\le \max_{x_t, x_{t-1:\eta+1}} p(x_t \mid x_{t-1:\eta+1}) \le \max_{x_t, x_{t-1:\eta}} p(x_t \mid x_{t-1:\eta}). \end{aligned} \tag{A.12}$$

Applying the inequalities in Eq. (A.12) term-by-term to $P_{\text{approx}}(\eta)$ (noting that any individual term in Eq. (A.12) satisfies $p(x_t \mid x_{t-1:\eta}) \le 1$) results in

$$
\begin{aligned}
&\min P_{\text{exact}} = \min P_{\text{approx}}(1) \le \min P_{\text{approx}}(2) \le \ldots \le \min P_{\text{approx}}(t-1) \\
&\le \max P_{\text{approx}}(t-1) \le \ldots \le \max P_{\text{approx}}(2) \le \max P_{\text{approx}}(1) = \max P_{\text{exact}},
\end{aligned}
\tag{A.13}
$$

which is Eq. (6.6). While Eq. (A.13) is exact its evaluation would be non-trivial except for short sequence lengths. For longer lengths, methods such as beam search [19] can only approximate $\max P_{\text{approx}}(\eta)$ and may violate these strict inequalities.

**Acknowledgement**

We thank Dr. Hongfeng Lou for the elegant proof of Eq. (A.1).